\documentclass[letterpaper]{article}
\usepackage{aaai19}  
\usepackage{times}  
\usepackage{helvet}  
\usepackage{courier}  
\usepackage{url}  
\usepackage{multirow}
\usepackage{booktabs}
\usepackage[pdftex]{graphicx}
\usepackage{multirow}
\usepackage{array}
\usepackage{amsfonts}
\usepackage{amsmath,bm}
\usepackage{enumitem}

\usepackage{xcolor}
\usepackage{soul}
\usepackage{mathtools}
\usepackage{subfigure}
\usepackage{amssymb}

\begin{document}


\title{A Deep Neural Network for Unsupervised Anomaly Detection and Diagnosis in Multivariate Time Series Data}


\author{Chuxu Zhang${}^{\S}$\thanks{This work was done when the first and fourth authors were summer interns at NEC Laboratories America. Dongjin Song is the corresponding author.}, Dongjin Song${}^{\dagger*}$, Yuncong Chen${}^{\dagger}$, Xinyang Feng${}^{\ddagger*}$, Cristian Lumezanu${}^{\dagger}$, \\\Large\textbf{Wei Cheng${}^{\dagger}$, Jingchao Ni${}^{\dagger}$, Bo Zong${}^{\dagger}$, Haifeng Chen${}^{\dagger}$, Nitesh V. Chawla${}^{\S}$}\\
${}^{\S}$University of Notre Dame, IN 46556, USA\\
${}^{\dagger}$NEC Laboratories America, Inc., NJ 08540, USA\\
${}^{\ddagger}$Columbia University, NY 10027, USA\\
${}^{\S}$\{czhang11,nchawla\}@nd.edu, ${}^{\dagger}$\{dsong,yuncong,lume,weicheng,jni,bzong,haifeng\}@nec-labs.com, ${}^{\ddagger}$xf2143@columbia.edu
}

\maketitle
\begin{abstract}
Nowadays, multivariate time series data are increasingly collected in various real world systems, \textit{e.g.}, power plants, wearable devices, \textit{etc}. Anomaly detection and diagnosis in multivariate time series refer to identifying abnormal status in certain time steps and pinpointing the root causes. Building such a system, however, is challenging since it not only requires to capture the temporal dependency in each time series, but also need encode the inter-correlations between different pairs of time series. In addition, the system should be robust to noise and provide operators with different levels of anomaly scores based upon the severity of different incidents. Despite the fact that a number of unsupervised anomaly detection algorithms have been developed, few of them can jointly address these challenges. In this paper, we propose a Multi-Scale Convolutional Recurrent Encoder-Decoder (MSCRED), to perform anomaly detection and diagnosis in multivariate time series data. Specifically, MSCRED first constructs multi-scale (resolution) signature matrices to characterize multiple levels of the system statuses in different time steps. Subsequently, given the signature matrices, a convolutional encoder is employed to encode the inter-sensor (time series) correlations and an attention based Convolutional Long-Short Term Memory (ConvLSTM) network is developed to capture the temporal patterns. Finally, based upon the feature maps which encode the inter-sensor correlations and temporal information, a convolutional decoder is used to reconstruct the input signature matrices and the residual signature matrices are further utilized to detect and diagnose anomalies. Extensive empirical studies based on a synthetic dataset and a real power plant dataset demonstrate that MSCRED can outperform state-of-the-art baseline methods.

\end{abstract}









\section{Introduction}

\begin{figure}
\begin{center}
\includegraphics[scale=0.41]{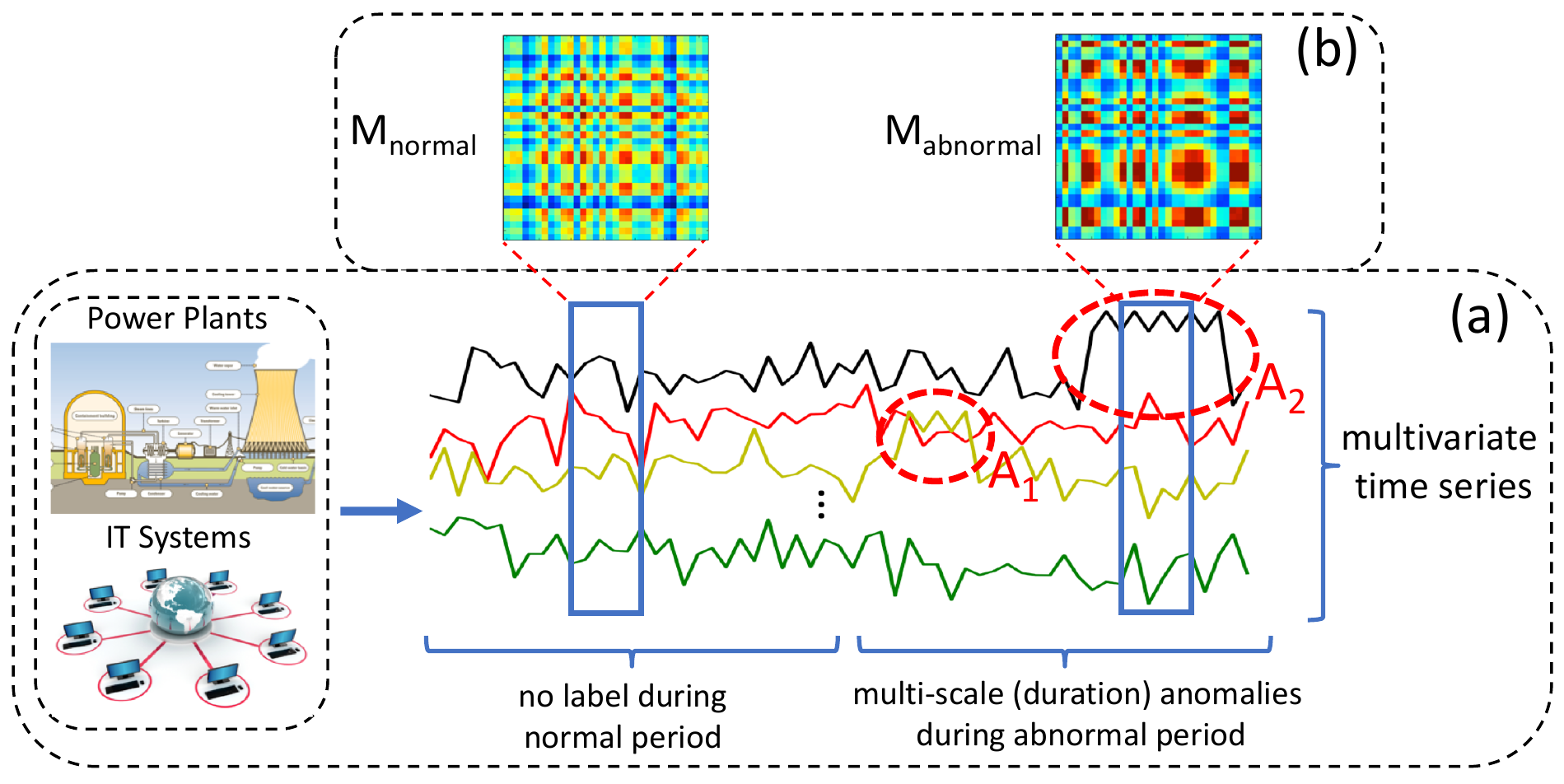}
\vspace{-0.15in}
\caption{(a) Unsupervised anomaly detection and diagnosis in multivariate time series data. (b) Different system signature matrices between normal and abnormal periods.}
\vspace{-0.25in}
\label{fig: challenge}
\end{center}
\end{figure}

Complex systems are ubiquitous in modern manufacturing industry and information services. Monitoring the behaviors of these systems generates a substantial amount of multivariate time series data, such as the readings of the networked sensors (\textit{e.g.}, temperature and pressure) distributed in a power plant or the connected components (\textit{e.g.}, CPU usage and disk I/O) in an Information Technology (IT) system. A critical task in managing these systems is to detect anomalies in certain time steps such that the operators can take further actions to resolve underlying issues. For instance, an anomaly score can be produced based on the sensor data and it can be used as an indicator of power plant failure \cite{len2007application}. An accurate detection is crucial to avoid serious financial and business losses as it has been reported that 1 minute downtime of an automotive manufacturing plant may cost up to $20,000$ US dollars \cite{djurdjanovic2003watchdog}. In addition, pinpointing the root causes, \textit{i.e.}, identifying which sensors (system components) are causes to an anomaly, can help the system operator perform system diagnosis and repair in a timely manner. In real world applications, it is common that a short term anomaly caused by temporal turbulence or system status switch may not eventually lead to a true system failure due to the auto-recovery capability and robustness of modern systems. Therefore, it would be ideal if an anomaly detection algorithm can provide operators with different levels of anomaly scores based upon the severity of various incidents. For simplicity, we assume that the severity of an incident is proportional to the duration of an anomaly in this work.
Figure \ref{fig: challenge}(a) illustrates two anomalies, \textit{i.e.}, $A_{1}$ and $A_{2}$ marked by red dash circle, in multivariate time series data. The root causes are yellow and black time series, respectively. The duration (severity level) of $A_{2}$ is larger than $A_{1}$.

To build a system which can automatically detect and diagnose anomalies, one main problem is that few or even no anomaly label is available in the historical data, which makes the supervised algorithms \cite{gornitz2013toward} infeasible. In the past few years, a substantial amount of unsupervised anomaly detection methods have been developed. The most prominent techniques include distance/clustering methods \cite{he2003discovering,ismo2004outlier,ide2007computing,campello2015hierarchical}, probabilistic methods \cite{chandola2009anomaly}, density estimation methods \cite{manevitz2001one}, temporal prediction approaches \cite{chen2008exploiting,gunnemann2014robust}, and the more recent deep learning techniques \cite{malhotra2016lstm,qin2017dual,zhou2017anomaly,wu2018restful,zong2018deep}. Despite the intrinsic unsupervised setting, most of them may still not be able to detect anomalies effectively due to the following reasons:

\begin{itemize}[leftmargin=0.15in]\setlength{\itemsep}{0pt}
\item There exists temporal dependency in multivariate time series data. Due to this reason, distance/clustering methods, \textit{e.g.}, k-Nearest Neighbor (kNN)~\cite{ismo2004outlier}), classification methods, \textit{e.g.}, One-Class SVM~\cite{manevitz2001one}, and density estimation methods, \textit{e.g.}, Deep Autoencoding Gaussian Mixture Model (DAGMM)~\cite{zong2018deep}, may not perform well since they cannot capture temporal dependencies across different time steps. 
\item Multivariate time series data usually contain noise in real word applications. When the noise becomes relatively severe, it may affect the generalization capability of temporal prediction models, \textit{e.g.}, Autoregressive Moving Average (ARMA) \cite{hamilton1994time} and LSTM encoder-decoder \cite{malhotra2016lstm,qin2017dual}, and increase the false positive detections. 
\item In real world application, it is meaningful to provide operators with different levels of anomaly scores based upon the severity of different incidents. The existing methods for root cause analysis, \textit{e.g.}, Ranking Causal Anomalies (RCA) \cite{cheng2016ranking}, are sensitive to noise and cannot handle this issue.
\end{itemize}

In this paper, we propose a Multi-Scale Convolutional Recurrent Encoder-Decoder (MSCRED) to jointly consider the aforementioned issues. Specifically, MSCRED first constructs multi-scale (resolution) signature matrices to characterize multiple levels of the system statuses across different time steps. In particular, different levels of the system statuses are used to indicate the severity of different abnormal incidents. Subsequently, given the signature matrices, a convolutional encoder is employed to encode the inter-sensor (time series) correlations patterns and an attention based Convolutional Long-Short Term Memory (ConvLSTM) network is developed to capture the temporal patterns. Finally, with the feature maps which encode the inter-sensor correlations and temporal information, a convolutional decoder is used to reconstruct the signature matrices and the residual signature matrices are further utilized to detect and diagnose anomalies. The intuition is that MSCRED may not reconstruct the signature matrices well if it never observes similar system statuses before. For example, Figure \ref{fig: challenge}(b) shows two signature matrices $M_{\textrm{normal}}$ and $M_{\textrm{abnormal}}$ during normal and abnormal periods. Ideally, MSCRED cannot reconstruct $M_{\textrm{abnormal}}$ well as training matrices (\textit{e.g.}, $M_{\textrm{normal}}$) are distinct from $M_{\textrm{abnormal}}$. To summarize, the main contributions of our work are:
\begin{itemize}[leftmargin=0.15in]\setlength{\itemsep}{0pt}
\item We formulate the anomaly detection and diagnosis problem as three underlying tasks, \textit{i.e.}, anomaly detection, root cause identification, and anomaly severity (duration) interpretation. Unlike previous studies which investigate each problem independently, we address these issues jointly. 
\item We introduce the concept of system signature matrix, develop MSCRED to encode the inter-sensor correlations via a convolutional encoder, incorporate temporal patterns with attention based ConvLSTM networks, and reconstruct signature matrix via a convolutional decoder. As far as we know, MSCRED is the first model that considers correlations among multivariate time series for anomaly detection and can jointly resolve all the three tasks.
\item We conduct extensive empirical studies on a synthetic dataset as well as a power plant dataset. Our results demonstrate the superior performance of MSCRED over state-of-the-art baseline methods. 
\end{itemize}

\section{Related Work}\label{sec:related}
Unsupervised anomaly detection on multivariate time series data is a challenging task and various types of approaches have been developed in the past few years. 

One traditional type is the distance methods~\cite{ismo2004outlier,ide2007computing}. For instance, the $k$-Nearest Neighbor (kNN) algorithm \cite{ismo2004outlier} computes the anomaly score of each data sample based on the average distance to its $k$ nearest neighbors. Similarly, the clustering models \cite{he2003discovering,campello2015hierarchical} cluster different data samples and find anomalies via a predefined outlierness score. In addition, the classification methods, \textit{e.g.}, One-Class SVM \cite{manevitz2001one}, models the density distribution of training data and classifies new data as normal or abnormal. Although these methods have demonstrated their effectiveness in various applications, they may not work well on multivariate time series since they cannot capture the temporal dependencies appropriately.
To address this issue, temporal prediction methods, \textit{e.g.}, Autoregressive Moving Average (ARMA) \cite{hamilton1994time} and its variants~\cite{brockwell2013time}, have been used to model temporal dependency and perform anomaly detection. However, these models are sensitive to noise and thus may increase false positive results when noise is severe. Other traditional methods include correlation methods \cite{kriegel2012outlier}, ensemble methods \cite{lazarevic2005feature}, \textit{etc.}


Besides traditional methods, deep learning based unsupervised anomaly detection algorithms \cite{malhotra2016lstm,zhai2016deep,zhou2017anomaly,zong2018deep} have gained a lot attention recently. For instance, Deep Autoencoding Gaussian Mixture Model (DAGMM) \cite{zong2018deep} jointly considers deep auto-encoder and Gaussian mixture model to model density distribution of multi-dimensional data. LSTM encoder-decoder \cite{malhotra2016lstm,qin2017dual} models time series temporal dependency by LSTM networks and achieves better generalization capability than traditional methods. 
Despite their effectiveness, they cannot jointly consider the temporal dependency, noise resistance, and the interpretation of severity of anomalies. 

In addition, our model design is inspired by fully convolutional neural networks \cite{long2015fully}, convolutional LSTM networks \cite{xingjian2015convolutional}, and attention technique \cite{bahdanau2014neural,yang2016hierarchical}. This paper is also related to other time series applications such as clustering/classification \cite{li2011time,hallac2017toeplitz,karim2018lstm}, segmentation \cite{keogh2001online,lemire2007better}, and so on. 

\section{MSCRED Framework}\label{sec:model}
In this section, we first introduce the problem we aim to study and then we elaborate the proposed Multi-Scale Convolutional Recurrent Encoder-Decoder (MSCRED) in detail. Specifically,
we first show how to generate multi-scale (resolution) system signature matrices. Then, we encode the spatial information in signature matrices via a convolutional encoder and model the temporal information via an attention based ConvLSTM. Finally, we reconstruct signature matrices based upon a convolutional decoder and use a square loss to perform end-to-end learning.

\subsection{Problem Statement}\label{sec:problem}
Given the historical data of $n$ time series with length $T$, \textit{i.e.}, $X = (\mathbf{x}_{1}, \cdots, \mathbf{x}_{n})^{T}\in\mathbb{R}^{n\times T}$, and assuming that there exists no anomaly in the data, we aim to achieve two goals: 
\begin{itemize}[leftmargin=0.15in]\setlength{\itemsep}{0pt}
\item \textbf{Anomaly detection}, \textit{i.e.}, detecting anomaly events at certain time steps after $T$.
\item \textbf{Anomaly diagnosis}, \textit{i.e.}, given the detection results, identifying the abnormal time series that are most likely to be the causes of each anomaly and interpreting the anomaly severity (duration scale) qualitatively. 
\end{itemize}
 
\subsection{Characterizing Status with Signature Matrices}
The previous studies \cite{hallac2017toeplitz,song2018deep} suggest that the correlations between different pairs of time series are critical to characterize the system status. To represent the inter-correlations between different pairs of time series in a multivariate time series segment from $t-w$ to $t$, we construct an $n \times n$ signature matrix $M^{t}$ based upon the pairwise inner-product of two time series within this segment. Two examples of signature matrices are shown in Figure \ref{fig: challenge}(b). Specifically, given two time series $\mathbf{x}_{i}^{w} = ({x}_{i}^{t-w}, {x}_{i}^{t-w-1},\cdots, {x}_{i}^{t})$ and $\mathbf{x}_{j}^{w} = ({x}_{j}^{t-w}, {x}_{j}^{t-w-1},\cdots, {x}_{j}^{t})$ in a multivariate time series segment $X^{w}$, their correlation $m_{ij}^{t} \in M^{t}$ is calculated with: 
\begin{equation}
\small
\begin{split}
\vspace{-0.1in}
m_{ij}^{t} = \frac{\sum _{\delta = 0}^{w} {x}_{i}^{t-\delta }{x}_{j}^{t-\delta }}{\kappa}
\vspace{-0.1in}
\end{split}
\label{equ: inner-product}
\end{equation}
where $\kappa$ is a rescale factor ($\kappa = w$). The signature matrix, \textit{i.e.}, $M^{t}$, not only can capture the shape similarities and value scale correlations between two time series, but also is robust to input noise as the turbulence at certain time series has little impact on the signature matrices. In this work, the interval between two segments is set as 10. In addition, to characterize system status at different scales, we construct $s$ ($s$ = 3) signature matrices with different lengths ($w$ = 10, 30, 60) at each time step. 

\subsection{Convolutional Encoder}
\begin{figure*}
\begin{center}
\includegraphics[scale=0.48]{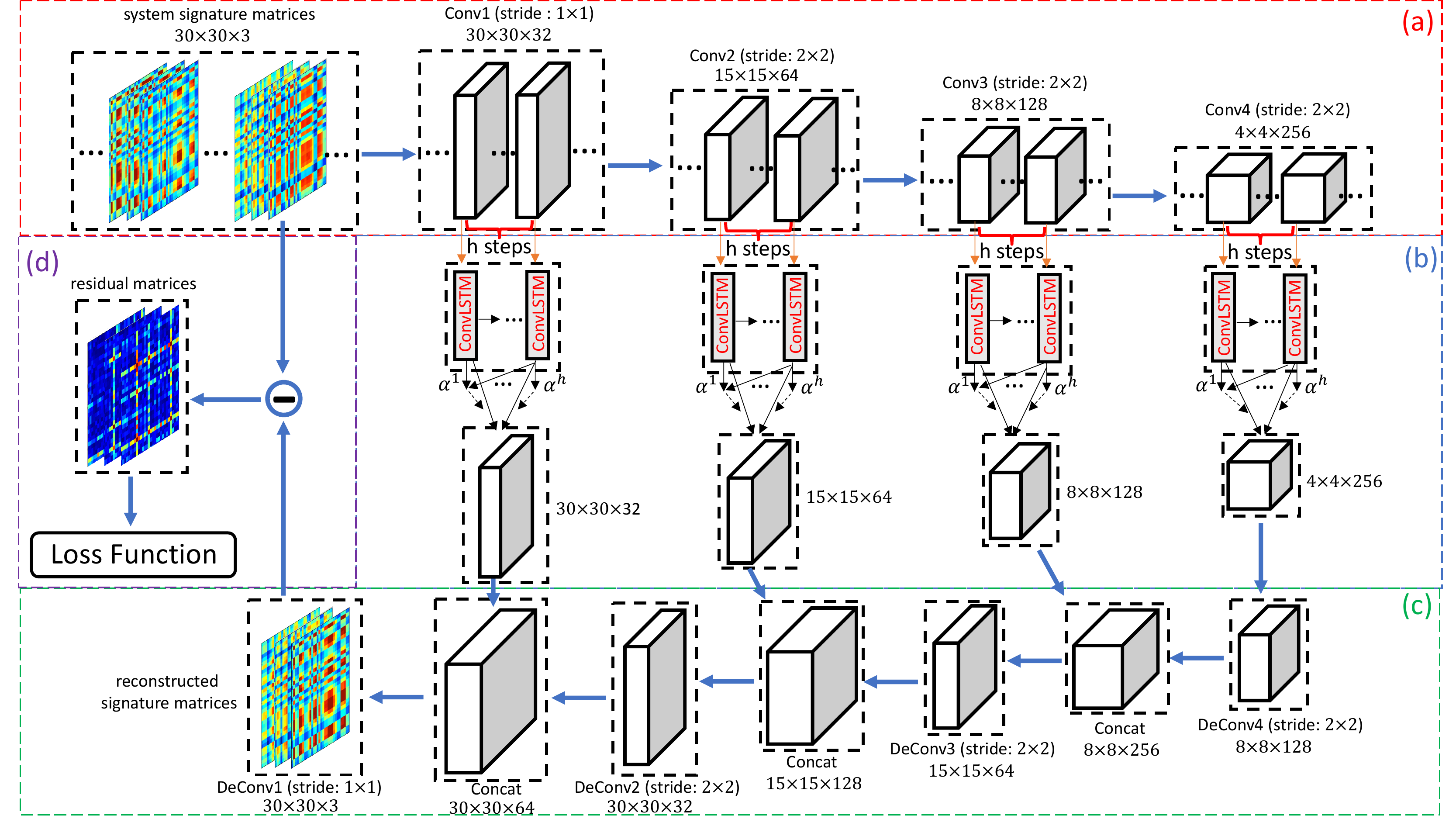}
\vspace{-0.15in}
\caption{Framework of the proposed model: (a) Signature matrices encoding via fully convolutional neural networks. (b) Temporal patterns modeling by attention based convolutional LSTM networks. (c) Signature matrices decoding via deconvolutional neural networks. (d) Loss function.}
\vspace{-0.2in}
\label{fig: model_illustration}
\end{center}
\end{figure*}

We employ a fully convolutional encoder \cite{long2015fully} to encode the spatial patterns of system signature matrices. Specifically, we concatenate $M^{t}$ at different scales as a tensor $\mathcal{X}^{t, 0}\in \mathbb{R}^{n \times n \times s}$, and then feed it to a number of convolutional layers. Assuming that $\mathcal{X}^{t, l-1}\in \mathbb{R}^{n_{l-1} \times n_{l-1} \times d_{l-1}}$ denotes the feature maps in the ($l-1$)-th layer, the output of $l$-th layer is given by:
\begin{equation}
\small
\begin{split}
\vspace{-0.1in}
\mathcal{X}^{t, l} = f( W^{l} \ast \mathcal{X}^{t, l-1} +b^{l})
\vspace{-0.1in}
\end{split}
\label{equ: CNN}
\end{equation}
where $\ast$ denotes the convolutional operation, $f(\cdot)$ is the activation function, $W^{l}\in \mathbb{R}^{k_{l}\times k_{l} \times d_{l-1}\times d_{l}}$ denotes $d_l$ convolutional kernels of size $k_{l}\times k_{l} \times d_{l-1}$, $b^{l}\in \mathbb{R}^{d_{l}}$ is a bias term, and $\mathcal{X}^{t, l}\in \mathbb{R}^{n_{l} \times n_{l} \times d_{l}}$ denotes the output feature map at $l$-th layer.
In this work, we use Scaled Exponential Linear Unit (SELU) \cite{klambauer2017self} as the activation function and 4 convolutional layers, \textit{i.e.}, Conv1-Conv4 with 32 kernels of size $3\times 3 \times 3$, 64 kernels of size $3\times 3  \times 32$, 128 kernels of size $2\times 2  \times 64$, and 256 kernels of size $2\times 2  \times 128$, as well as $1\times 1$, $2 \times 2$, $2 \times 2$, and $2 \times 2$ strides, respectively. Note that the exact order of the time series based on which the signature matrices are formed is not important, because for any given permutation, the resulting local patterns can be captured by the convolutional encoder.
Figure \ref{fig: model_illustration}(a) illustrates the detailed encoding process of signature matrices.

\subsection{Attention based ConvLSTM}
The spatial feature maps generated by convolutional encoder is temporally dependent on previous time steps. Although ConvLSTM~\cite{xingjian2015convolutional} has been developed to capture the temporal information in a video sequence, its performance may deteriorate as the length of sequence increases. To address this issue, we develop an attention based ConvLSTM which can adaptively select relevant hidden states (feature maps) across different time steps. Specifically, given the feature maps $\mathcal{X}^{t, l}$ from the $l$-th convolutional layer and previous hidden state $\mathcal{H}^{t-1, l}\in \mathbb{R}^{n_{l} \times n_{l} \times d_{l}}$, the current hidden state $\mathcal{H}^{t, l}$ is updated with $\mathcal{H}^{t, l} = \textrm{ConvLSTM}(\mathcal{X}^{t, l}, \mathcal{H}^{t-1, l})$, where the ConvLSTM cell~\cite{xingjian2015convolutional} is formulated as:  
\begin{equation}
\small
\begin{split}
\vspace{-0.1in}
&{\bf z}^{t, l} = \sigma (\tilde{W}_{\mathcal{X}\mathcal{Z}}^{l}\ast \mathcal{X}^{t, l} + \tilde{W}_{\mathcal{H}\mathcal{Z}}^{l}\ast \mathcal{H}^{t-1, l} + \tilde{W}_{\mathcal{C}\mathcal{Z}}^{k}\circ \mathcal{C}^{t-1, l} + \tilde{b}_{\mathcal{Z}}^{l}) \\
&{\bf r}^{t, l} = \sigma (\tilde{W}_{\mathcal{X}\mathcal{R}}^{l}\ast \mathcal{X}^{t, l} + \tilde{W}_{\mathcal{H}\mathcal{R}}^{l}\ast \mathcal{H}^{t-1, l} + \tilde{W}_{\mathcal{C}\mathcal{R}}^{l}\circ \mathcal{C}^{t-1, l} + \tilde{b}_{\mathcal{R}}^{l}) \\
& \mathcal{C}^{t, l} = {\bf z}^{t, l}\circ \tanh(\tilde{W}_{\mathcal{X}\mathcal{C}}^{l}\ast \mathcal{X}^{t, l} + \tilde{W}_{\mathcal{H}\mathcal{C}}^{l}\ast \mathcal{H}^{t-1, l} + \tilde{b}_{\mathcal{C}}^{l}) + \\
&~~~~~~~~~~~~{\bf r}^{t, l}\circ \mathcal{C}^{t-1, l}\\
&{\bf o}^{t, l} = \sigma (\tilde{W}_{\mathcal{X}\mathcal{O}}^{l}\ast \mathcal{X}^{t, l} + \tilde{W}_{\mathcal{H}\mathcal{O}}\ast \mathcal{H}^{t-1, l} + \tilde{W}_{\mathcal{C}\mathcal{O}}\circ \mathcal{C}^{t, l} + \tilde{b}_{\mathcal{O}}^{l}) \\
& \mathcal{H}^{t, l} = {\bf o}^{t, l}\circ \tanh(\mathcal{C}^{t, l})
\vspace{-0.1in}
\end{split}
\label{equ: convlstm}
\end{equation}
where $\ast$ denotes the convolutional operator, $\circ$ represents Hadamard product, $\sigma$ is the sigmoid function, $\tilde{W}^{l}_{\mathcal{X}\mathcal{Z}},\tilde{W}^{l}_{\mathcal{H}\mathcal{Z}}, \tilde{W}^{l}_{\mathcal{C}\mathcal{Z}}, \tilde{W}^{l}_{\mathcal{X}\mathcal{R}}, \tilde{W}^{l}_{\mathcal{H}\mathcal{R}}, \tilde{W}^{l}_{\mathcal{C}\mathcal{R}},\tilde{W}^{l}_{\mathcal{X}\mathcal{C}}, \tilde{W}^{l}_{\mathcal{H}\mathcal{C}},
\tilde{W}^{l}_{\mathcal{X}\mathcal{O}},$
$\tilde{W}^{l}_{\mathcal{H}\mathcal{O}}, \tilde{W}^{l}_{\mathcal{C}\mathcal{O}}\in \mathbb{R}^{\tilde{k}_{l}\times \tilde{k}_{l} \times \tilde{d}_{l}\times \tilde{d}_{l}}$ are $\tilde{d}_{l}$ convolutional kernels of size $\tilde{k}_{l}\times \tilde{k}_{l} \times \tilde{d}_{l}$ and $\tilde{b}_{\mathcal{Z}}^{l},\tilde{b}_{\mathcal{R}}^{l},\tilde{b}_{\mathcal{C}}^{l},\tilde{b}_{\mathcal{O}}^{l}\in \mathbb{R}^{\tilde{d}_{l}}$ are bias parameters of the $l$-th layer ConvLSTM. In our work, we maintain the same convolutional kernel size as convolutional encoder at each layer. Note that all the input $\mathcal{X}^{t, l}$, cell outputs $\mathcal{C}^{t, l}$, hidden states $\mathcal{H}^{t-1, l}$, and gates ${\bf z}^{t, l}$, ${\bf r}^{t, l}$, ${\bf o}^{t, l}$ are 3D tensors, which is different from LSTM. We tune step length $h$ (\textit{i.e.,} the number of previous segments) and set it as 5 due to the best empirical performance. In addition, considering not all previous steps are equally correlated to the current state $\mathcal{H}^{t, l}$, we adopt a temporal attention mechanism to adaptively select the steps that are relevant to current step and aggregate the representations of those informative feature maps to form a refined output of feature maps $\hat{\mathcal{H}}^{t, l}$, which is given by:
\begin{equation}
\small
\begin{split}
\vspace{-0.1in}
\hat{\mathcal{H}}^{t, l} = \sum_{i \in (t-h, t)} \alpha^{i} \mathcal{H}^{i, l},\alpha^{i} = \frac{\exp\{\frac{\textrm{Vec}(\mathcal{H}^{t, l})^{\textrm{T}} \textrm{Vec}(\mathcal{H}^{i, l})}{\chi}\}}{\sum_{i \in (t-h, t)}\exp\{\frac{\textrm{Vec}(\mathcal{H}^{t, l})^{\textrm{T}}\textrm{Vec}(\mathcal{H}^{i, l})}{\chi}\}}
\vspace{-0.1in}
\end{split}
\label{equ: attention}
\end{equation}
where $\textrm{Vec}(\cdot)$ denotes vector and $\chi$ is a rescale factor ($\chi$ = 5.0). That is, we take the last hidden state $\mathcal{H}^{t, l}$ as the group level context vector and measure the importance weights $\alpha^{i}$ of previous steps through a softmax function. Unlike the general attention mechanism~\cite{bahdanau2014neural} that introduces transformation and context parameters, the above formulation is purely based on the learned hidden feature maps and achieves the similar function as the former. Essentially, the attention based ConvLSTM jointly models the spatial patterns of signature matrices with temporal information at each convolutional layer. Figure \ref{fig: model_illustration}(b) illustrates the temporal modeling procedure. 

\subsection{Convolutional Decoder}
To decode the feature maps obtained in previous step and get the reconstructed signature matrices, we design a convolutional decoder which is formulated as: 
\begin{equation}
\small
\begin{split}
\vspace{-0.1in}
\hat{\mathcal{X}}^{t, l-1}  = \left\{\begin{matrix}
 f({\hat{W}}^{t, l} \circledast \hat{\mathcal{H}}^{t, l}  + {\hat{b}}^{t, l}),& l = 4\\ 
 f({\hat{W}}^{t, l} \circledast [\hat{\mathcal{H}}^{t, l}\oplus \hat{\mathcal{X}}^{t, l}]  + {\hat{b}}^{t, l}) ,& l = 3, 2, 1
\end{matrix}\right.
\vspace{-0.1in}
\end{split}
\label{equ: DeCNN}
\end{equation}
where $\circledast$ denotes the deconvolution operation, $\oplus$ is the concatenation operation, $f(\cdot)$ is the activation unit (same as the encoder), ${\hat{W}}^{l} \in \mathbb{R}^{\hat{k}_{l}\times \hat{k}_{l} \times \hat{d}_{l} \times \hat{d}_{l-1}}$ and ${\hat{b}}^{l}\in \mathbb{R}^{\hat{d}_{l}}$ are filter kernel and bias parameter of $l$-th deconvolutional layer. Specifically, we follow the reverse order and feed $\hat{\mathcal{H}}^{t, l}$ of $l$-th ConvLSTM layer to a deconvolutional neural network. The output feature map $\hat{\mathcal{X}}^{t, l-1}\in \mathbb{R}^{\hat{n}_{l-1} \times \hat{n}_{l-1}\times \hat{d}_{l-1}}$ is concatenated with the output of previous ConvLSTM layer, making the decoder process stacked. The concatenated representation is further fed into the next deconvolutional layer. The final output $\hat{\mathcal{X}}^{t, 0}\in \mathbb{R}^{n \times n \times s}$ (with the same size of the input matrices) denotes the representations of reconstructed signature matrices. As a result, we use 4 deconvolutional layers: DeConv4-DeConv1 with 128 kernels of size $2\times 2 \times 256$, 64 kernels of size $2\times 2 \times 128$, 32 kernels of size $3\times 3 \times 64$, and 3 kernels of size $3\times 3 \times 64$ filters, as well as $2\times 2$, $2 \times 2$, $2 \times 2$, and $1 \times 1$ strides, respectively. The decoder is able to incorporate feature maps at different deconvolutional and ConvLSTM layers, which is effective to improve anomaly detection performance, as we will demonstrate in the experiment. Figure \ref{fig: model_illustration}(c) illustrates the decoding procedure. 

\subsection{Loss Function}
For MSCRED, the objective is defined as the reconstruction errors over the signature matrices, \textit{i.e.}, 
\begin{equation}
\small
\begin{split}
\vspace{-0.1in}
\mathcal{L}_{MSCRED} = \sum _{t}\sum_{c=1}^{s} \left \|\mathcal{X}^{t, 0}_{:,:,c} - \hat{\mathcal{X}}_{:,:,c}^{t, 0} \right \|^{2}_{F}
\vspace{-0.1in}
\end{split}
\label{equ: loss}
\end{equation}
where $\mathcal{X}^{t, 0}_{:,:,c}\in \mathbb{R}^{n\times n}$.
We employ mini-batch stochastic gradient descent method together with the Adam optimizer \cite{kingma2014adam} to minimize the above loss. After sufficient number of training epochs, the learned neural network parameters are utilized to infer the reconstructed signature matrices of validation and test data. Finally, we perform anomaly detection and diagnosis based on the residual signature matrices, which will be elaborated in the next section. 

\section{Experiments}\label{sec:exp}
In this section, we conduct extensive experiments to answer the following research questions:
\begin{itemize}[leftmargin=0.15in]\setlength{\itemsep}{0pt}
\item \textbf{Anomaly detection.} Whether MSCRED can outperform baseline methods for anomaly detection in multivariate time series (RQ1)? How does each component of MSCRED affect its performance (RQ2)?
\item \textbf{Anomaly diagnosis.} Whether MSCRED can perform root cause identification (RQ3) and anomaly severity (duration) interpretation (RQ4) effectively? 
\item \textbf{Robustness to noise.} Compared with baseline methods, whether MSCRED is more robust to input noise (RQ5)?
\end{itemize}

\subsection{Experimental Setup}
\subsubsection{Data.}
We use a synthetic dataset and a real world power plant dataset for empirical studies. The detailed statistics and settings of these two datasets are shown in Table 1. 
\begin{itemize}[leftmargin=0.15in]\setlength{\itemsep}{0pt}
\item \textbf{Synthetic data.}
Each time series is formulated as:
\begin{equation}
\small
\begin{split}
\vspace{-0.25in}
\mathcal{S}(t) = \left\{\begin{matrix}
 \underbrace{sin}_{C1}\underbrace{\left [(t - t_{0})/\omega \right ]}_{C2} + \underbrace{\lambda \cdot \epsilon}_{C3},& s_{\textrm{rand}} = 0 \\ 
 \underbrace{cos}_{C1}\underbrace{\left [(t - t_{0})/\omega \right ]}_{C2} + \underbrace{\lambda \cdot \epsilon}_{C3},& s_{\textrm{rand}} = 1
\end{matrix}\right. 
\vspace{-0.25in}
\end{split}
\label{equ: syn-data}
\end{equation}
where $s_{\textrm{rand}}$ is a 0 or 1 random seed. The above formula captures three attributes of multivariate time series: (a) trigonometric function (C1) simulates temporal patterns; (b) time delay $t_{0} \in [50, 100]$ and frequency $\omega \in [40, 50]$ (C2) simulates various periodic cycles; (c) random Gaussian noise $\epsilon \sim\mathcal{N}(0, 1)$ scaled by factor $\lambda = 0.3$ (C3) simulates data noise as well as various shapes. In addition, two sinusoidal waves have high correlation if their frequencies are similar and they are almost in-phase. By randomly selecting frequency and phase of each time series, we expect some pairs to have high correlations while some have low correlations. We randomly generate 30 time series and each includes 20000 points. Besides, 5 shock wave like anomalies (with similar value range of normal data, as the examples in Figure \ref{fig: challenge}(a)) are randomly injected into 3 random time series (root causes) during test period. The duration of each anomaly belongs to one of the three scales, \textit{i.e.}, 30, 60, 90. 
\item \textbf{Power plant data.}
This dataset was collected on a real power plant. It contains 36 time series generated by sensors distributed in the power plant system. It has 23,040 time steps and contains one anomaly identified by the system operator. Besides, we randomly inject 4 additional anomalies (similar to what we did in the synthetic data) into the test period for thorough evaluation. 
\end{itemize}

\begin{table}[!tb]
\begin{center}
\footnotesize
\caption{The detailed statistics and settings of two datasets.}
 \label{tab: data}
\begin{tabular}{c||c|c}
  \toprule
  {\bf Statistics}& {\bf Synthetic}&{\bf Power Plant}\\
   \midrule
    \# time series & 30 & 36 \\
   \# points & 20,000 & 23,040 \\
   \# anomalies & 5 & 5 \\
   \# root causes & 3 & 3 \\
   train period & 0 $\sim$ 8,000 & 0 $\sim$ 10,080\\
   valid period & 8,001 $\sim$ 10,000& 10,081 $\sim$ 18,720\\
   test period &10,001 $\sim$ 20,000 & 18,721 $\sim$ 23,040\\
    \midrule
  \end{tabular}     
  \vspace{-0.3in}
\end{center}
\end{table}%

\subsubsection{Baseline methods.}
We compare MSCRED with eight baseline methods of four categories, \textit{i.e.}, classification model, density estimation model, temporal prediction model, and variants of MSCRED.
\begin{itemize}[leftmargin=0.15in]\setlength{\itemsep}{0pt}
\item {\bf Classification model.} It learns a decision function and classifies test data as similar or dissimilar to the training set. We use One-Class SVM model (OC-SVM) \cite{manevitz2001one} for comparison.
\item {\bf Density estimation model.} It models data density for outlier detection. We use Deep Autoencoding Gaussian Mixture model (DAGMM) \cite{zong2018deep} and take the energy score \cite{zong2018deep} as the anomaly score. 
\item {\bf Prediction model.} It models the temporal dependencies of training data and predicts the value of test data. We employ three methods: History Average (HA), Auto-Regression Moving Average (ARMA) \cite{hamilton1994time} and LSTM encoder-decoder (LSTM-ED) \cite{cho2014learning}. The anomaly score is defined as the average prediction error over all time series. 
\item {\bf MSCRED variants.} Besides the above baseline methods, we consider three variants of MSCRED to justify the effectiveness of each component: (1) CNN$_{\textrm{ConvLSTM}}^{\textrm{ED}(4)}$ is MSCRED with attention module and first three ConvLSTM layers been removed. (2) CNN$_{\textrm{ConvLSTM}}^{\textrm{ED}(3,4)}$ is MSCRED with attention module and first two ConvLSTM layers been removed. (3) CNN$_{\textrm{ConvLSTM}}^{\textrm{ED}}$ is MSCRED with attention module been removed. 
\end{itemize}
We employ Tensorflow to implement MSCRED and its variants, and train them on a server with Intel(R) Xeon(R) CPU E5-2637 v4 3.50GHz and 4 NVIDIA GTX 1080 Ti graphics cards. The parameter settings of MSCRED are described in the model section. In addition, the anomaly score is defined as the number of poorly reconstructed pairwise correlations. In other words, the number of elements whose value is larger than a given threshold $\theta$ in the residual signature matrices and $\theta$ is detemined empirically over different datasets. 
\subsubsection{Evaluation metrics.}
We use three metrics, \textit{i.e.}, {\bf Precision}, {\bf Recall}, and {\bf F1 Score}, to evaluate the anomaly detection performance of each method. To detect anomaly, we follow the suggestion of a domain expert by setting a threshold $\tau = \beta \cdot \max\left \{ s(t)_{\textrm{valid}}\right \}$, where $s(t)_{\textrm{valid}}$ are the anomaly scores over the validation period and $\beta \in [1, 2]$ is set to maximize the F1 Score over the validation period. Recall and Precision scores over the test period are computed based on this threshold. 
Experiments on both datasets are repeated 5 times and the average results are reported for comparison. Note that the output of MSCRED contains three channel of residual signature matrices  \textit{w.r.t.} different segment lengths. We use the smallest one ($w$ = 10) for the following anomaly detection and root cause identification evaluation. The performance comparison of three channel results will also be provided for anomaly severity interpretation. 

\subsection{Performance Evaluation}
\subsubsection{Anomaly detection result (RQ1, RQ2).}
\begin{figure}
\begin{center}
\includegraphics[scale=0.35]{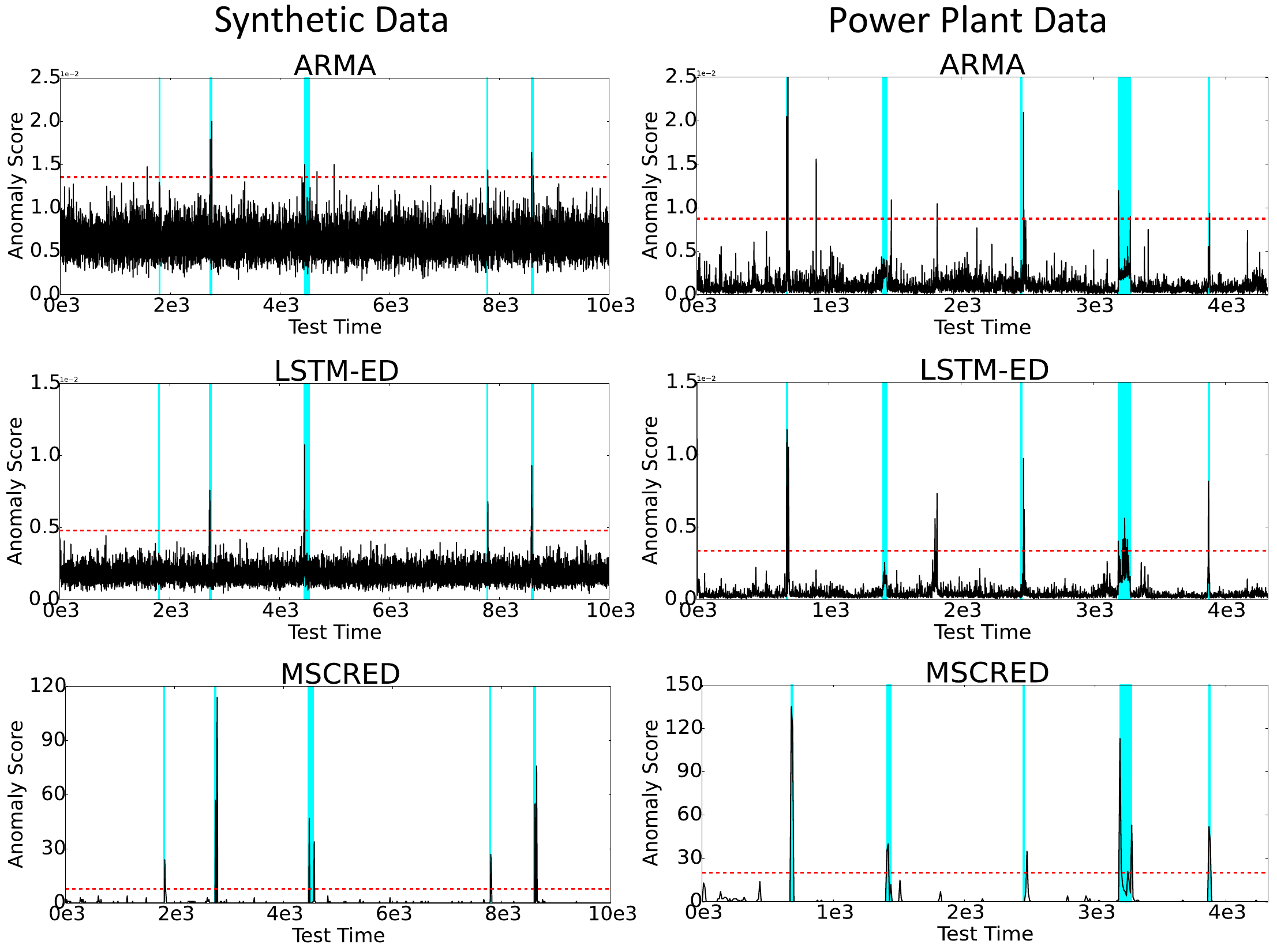}
\vspace{-0.15in}
\caption{Case study of anomaly detection. The shaded regions represent anomaly periods. The red dash line is the cutting threshold of anomaly.}
\vspace{-0.3in}
\label{fig: case-study-anomaly}
\end{center}
\end{figure}

\begin{figure}
\begin{center}
\includegraphics[scale=0.31]{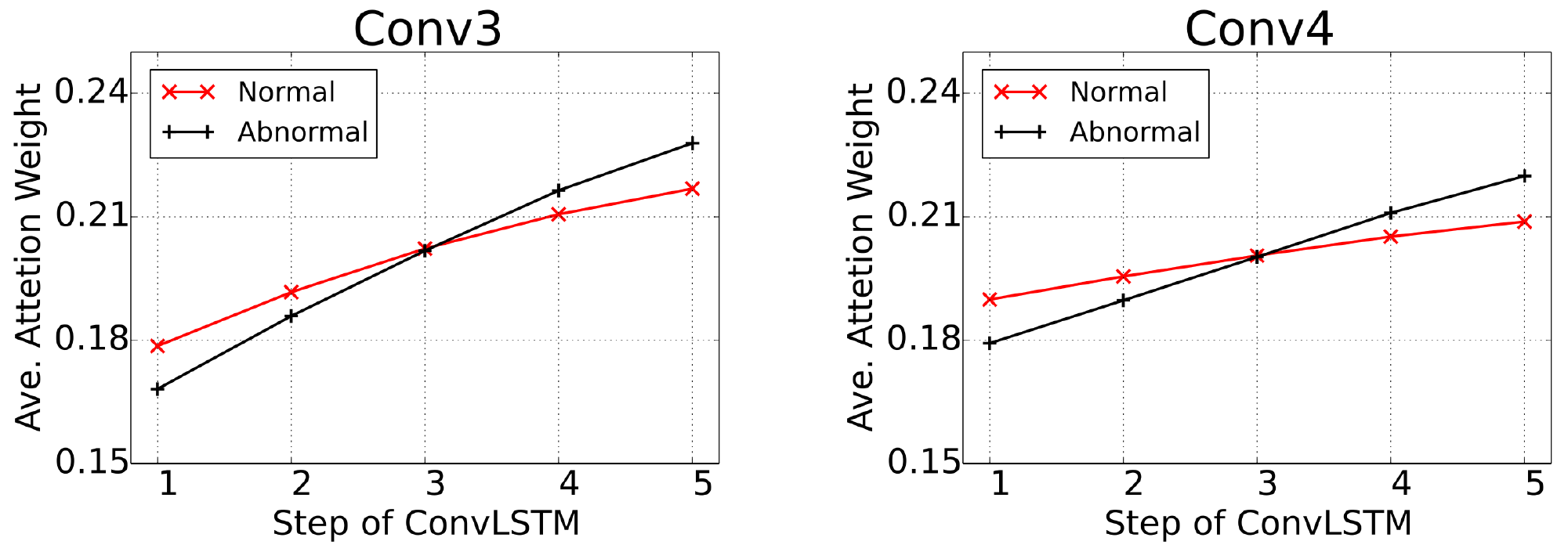}
\vspace{-0.15in}
\caption{Average distribution of attention weights at the last two ConvLSTM layers in the power plant data.}
\vspace{-0.2in}
\label{fig: attention_weight}
\end{center}
\end{figure}

The performance of different methods for anomaly detection are reported in Table \ref{tab: anomaly-result}, where the best scores are highlighted in bold-face and the best baseline scores are indicated by underline. The last row reports the improvement (\%) of MSCRED over the best baseline method. 
\begin{itemize}[leftmargin=0.15in]\setlength{\itemsep}{0pt}
\item {\bf (RQ1: comparison with baselines)} In Table \ref{tab: anomaly-result}, we observe that (a) temporal prediction models perform better than classification and density estimation models, indicating both datasets have temporal dependency; (b) LSTM-ED has better performance than ARMA, showing deep learning model can capture more complex relationship in the data than traditional method; (c) MSCRED performs best on all settings. The improvements over the best baseline range from 13.3\% to 30.0\%. In other words, MSCRED is much better than baseline methods as it can model both inter-sensor correlations and temporal patterns of multivariate time series effectively. 

In order to show the comparison in detail, Figure \ref{fig: case-study-anomaly} provides case study of MSCRED and two best baseline methods, \textit{i.e.}, ARMA and LSTM-ED, for both datasets. We can observe that the anomaly score of ARMA is not stable and the results contain many false positives and false negatives. Meanwhile, the anomaly score of LSTM-ED is smoother than ARMA while still contains several false positives and false negatives. MSCRED can detect all anomalies without any false positive and false negative. 

To demonstrate a more convincing evaluation, we do experiment on another synthetic data with 10 anomalies (it is easy to generate larger data with more anomalies). The average recall and precision scores (5 repeated experiments) of MSCRED are (0.84, 0.95) while the values of LSTM-ED are (0.64, 0.87). In addition, we do experiment on another large power plant data which has 920 sensors and 11 labeled anomalies. The recall and precision scores of MSCRED are (7/11, 7/13) while the values of LSTM-ED are (5/11, 5/17). All evaluation results show the effectiveness of our model.
\begin{table}[!tb]
\begin{center}
\footnotesize
\caption{Anomaly detection results on two datasets.}
\vspace{-0.1in}
 \label{tab: anomaly-result}
\begin{tabular}{c||c|c|c|c|c|c}
  \toprule
  \multirow{2}{*}{\bf Method}& \multicolumn{3}{c|}{\bf Synthetic Data}&\multicolumn{3}{c}{\bf Power Plant Data}\\
   \cline{2-7}
      & Pre & Rec & F$_{1}$ & Pre & Rec & F$_{1}$ \\
   \midrule
   \midrule
     OC-SVM & 0.14 & 0.44 & 0.22 &0.11  & 0.28 & 0.16 \\
      \midrule
     DAGMM & 0.33 & 0.20 & 0.25 & 0.26 & 0.20  & 0.23  \\
     \midrule
     HA & 0.71 & 0.52 & 0.60 & 0.48 & 0.52 & 0.50 \\
     ARMA & 0.91 & 0.52 & 0.66 & 0.58 & 0.60 & 0.59 \\
     LSTM-ED &\underline{1.00}  & \underline{0.56} & \underline{0.72} & \underline{0.75} & \underline{0.68} & \underline{0.71} \\
     \midrule
     CNN$_{ConvLSTM}^{ED(4)}$ &  0.37& 0.24 & 0.29 & 0.67 &0.56  & 0.61 \\
     CNN$_{ConvLSTM}^{ED(3,4)}$ &0.63  & 0.56 &0.59 & 0.80 & 0.72 &0.76  \\
    CNN$_{ConvLSTM}^{ED}$ & 0.80 &  0.76&  0.78& 0.85 & 0.72 & 0.78 \\
      \midrule
    MSCRED & {\bf 1.00} & {\bf 0.80} & {\bf 0.89} & {\bf 0.85} & {\bf 0.80} & {\bf 0.82} \\
     \midrule
    Gain (\%)& -- & 30.0 & 23.8 & 13.3 & 19.4 & 15.5 \\
    \midrule
  \end{tabular}     
  \vspace{-0.2in}
\end{center}
\end{table}%


\item {\bf (RQ2: comparison with model variants)} In Table \ref{tab: anomaly-result}, we also observe that by increasing the number of ConvLSTM layers, the performance of MSCRED improves. Specifically, CNN$_{\textrm{ConvLSTM}}^{\textrm{ED}}$ outperforms CNN$_{\textrm{ConvLSTM}}^{\textrm{ED}(3,4)}$ and the performance of CNN$_{\textrm{ConvLSTM}}^{\textrm{ED}(3,4)}$ is superior than CNN$_{\textrm{ConvLSTM}}^{\textrm{ED}(4)}$, indicating the effectiveness of ConvLSTM layers and stacked decoding process for model refinement. We also observe that CNN$_{\textrm{ConvLSTM}}^{\textrm{ED}}$ is worse than MSCRED, suggesting that attention based ConvLSTM can further improve anomaly detection performance. 

To further demonstrate the effectiveness of attention module, Figure \ref{fig: attention_weight} reports the average distribution of attention weights over 5 previous timesteps at the last two ConvLSTM layers. The results are obtained using the power plant data. We compute the average attention weights distribution for segments in the normal periods and that for segments in the abnormal periods separately. Note that in the latter distribution, the older timesteps (step 1 or 2), which tend to still be normal and therefore in a different system status than the current timestep (step 5), are assigned lower weights than in the distribution for normal segments. In other words, the attention modules show high sensitivity to system status change and thus is beneficial for anomaly detection.

\end{itemize}

\subsubsection{Root cause identification result (RQ3).}
\begin{figure}
\begin{center}
\includegraphics[scale=0.48]{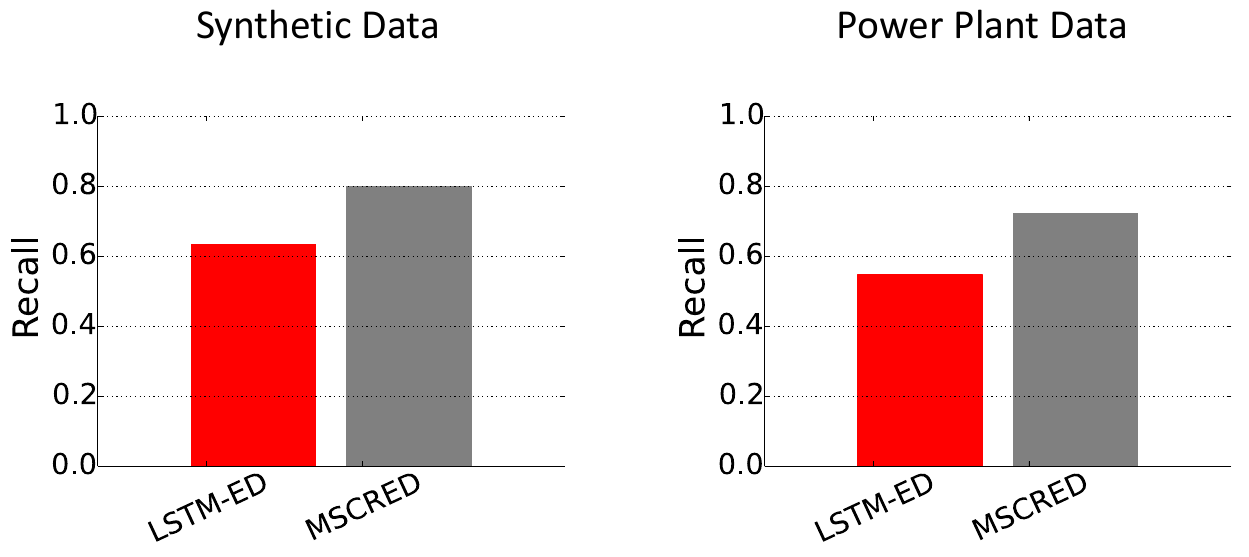}
\vspace{-0.2in}
\caption{Performance of root cause identification.}
\vspace{-0.3in}
\label{fig: root_cause_identification}
\end{center}
\end{figure}

As one of the anomaly diagnosis tasks, root cause identification depends on good anomaly detection performance. Therefore, we compare the performances of MSCRED and the best baseline, \textit{i.e.}, LSTM-ED. Specifically, for LSTM-ED, we use the prediction error of each time series to represent its anomaly score of this series. The same value of MSCRED is defined as the number of poorly reconstructed pairwise correlations in a specific row/column of residual signature matrices as each row/column denotes a time series. For each anomaly event, we rank all time series by their anomaly scores and identify the top-$k$ series as the root causes. Figure \ref{fig: root_cause_identification} shows the average recall$@k$ ($k$ = 3) in 5 repeated experiments. MSCRED outperforms LSTM-ED by a margin of 25.9\% and 32.4\% in the synthetic and power plant data, respectively.

\begin{figure}
\begin{center}
\includegraphics[scale=0.40]{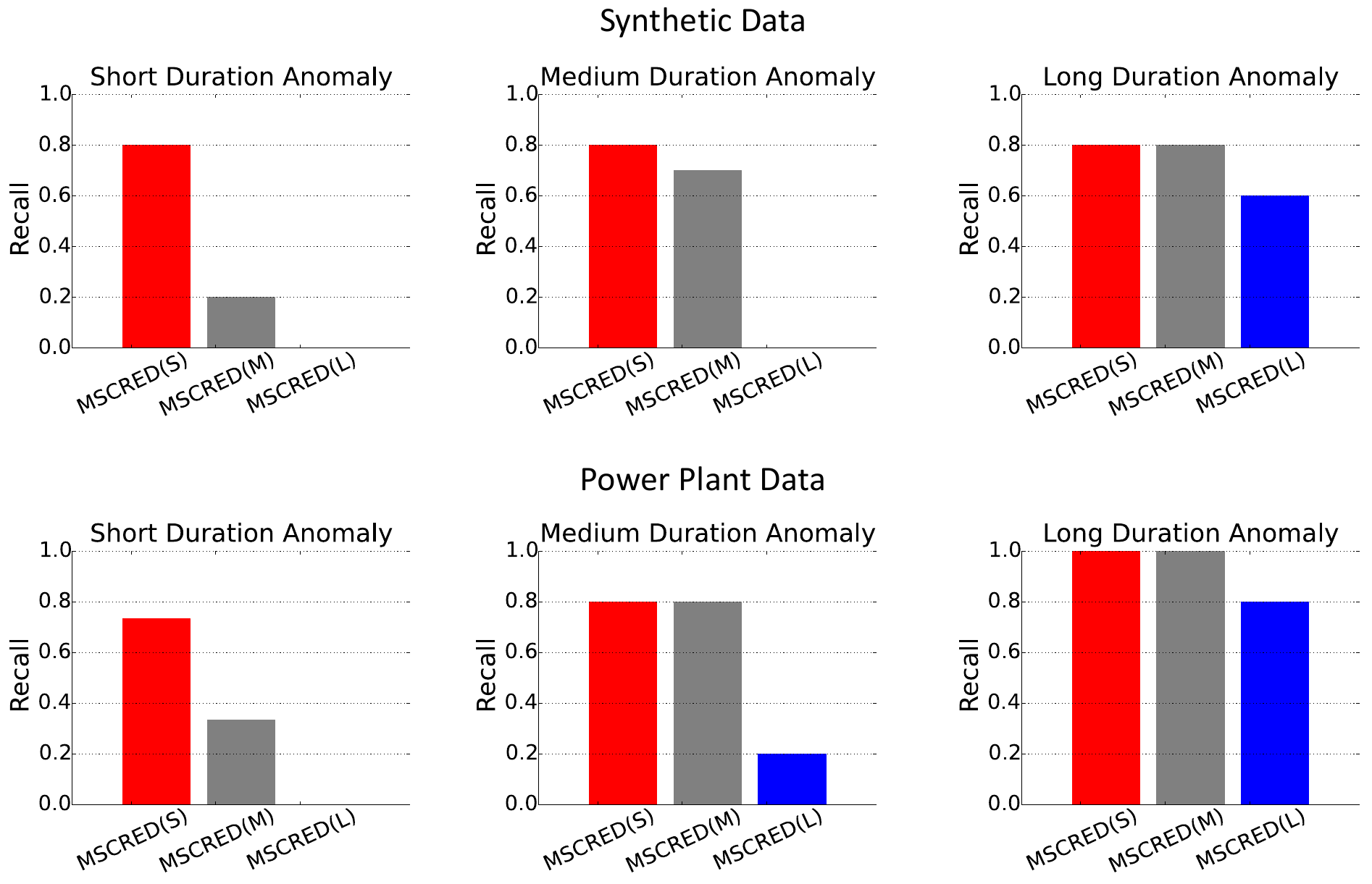}
\vspace{-0.15in}
\caption{Performance of three channels of MSCRED over different types of anomalies.}
\vspace{-0.3in}
\label{fig: anomaly_scale_analysis}
\end{center}
\end{figure}
\begin{figure}
\begin{center}
\includegraphics[scale=0.61]{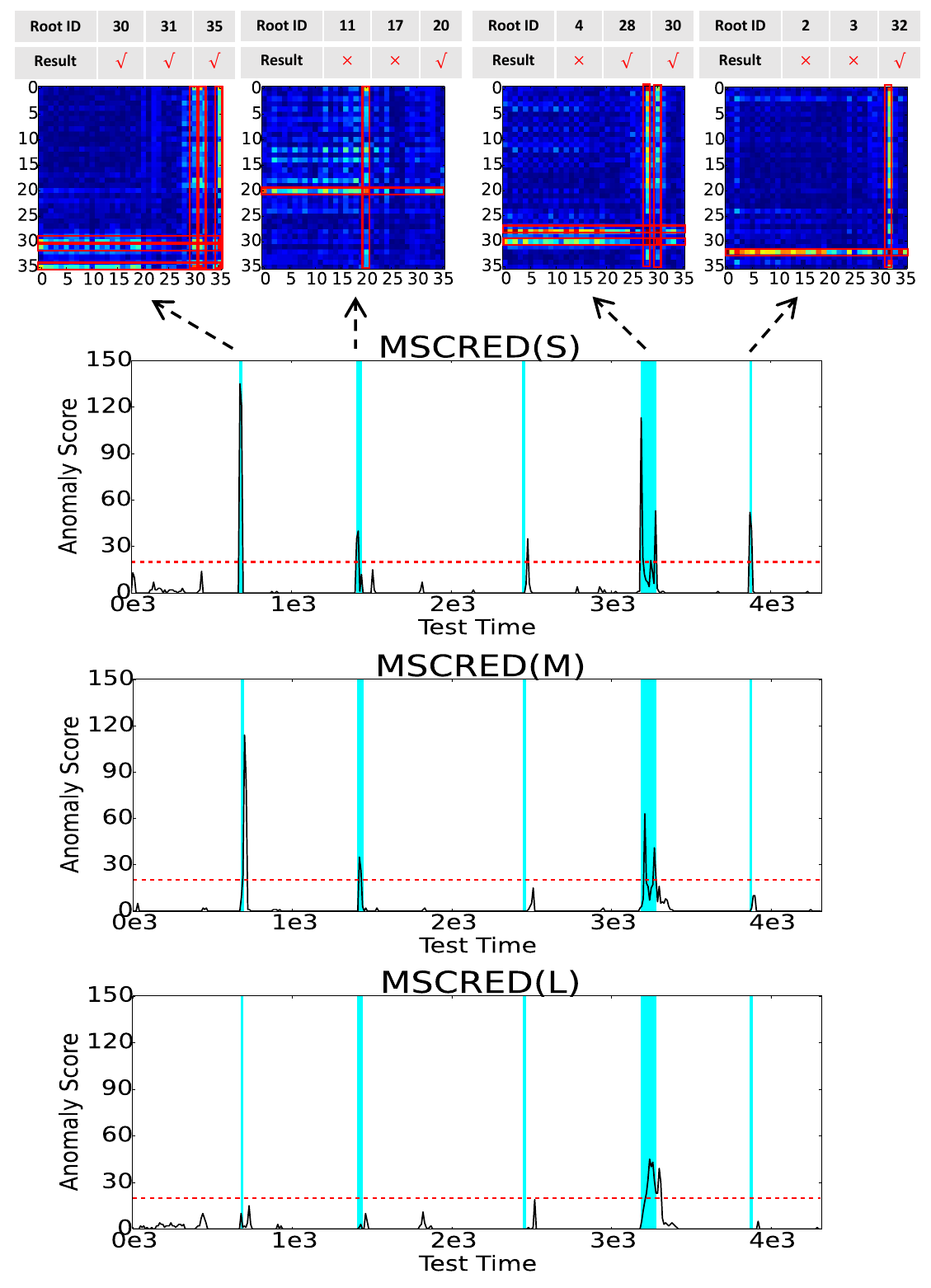}
\vspace{-0.2in}
\caption{Case study of anomaly diagnosis.}
\vspace{-0.2in}
\label{fig: diagnosis-case-study-MCRNN}
\end{center}
\end{figure}
\subsubsection{Anomaly severity (duration) interpretation (RQ4).}
The signature matrices of MSCRED include $s$ channels ($s$ = 3 in current experiments) that capture system status at different scales. To interpret anomaly severity, we first compute different anomaly scores based on the residual signature matrices of three channels, \textit{i.e.}, small, medium, and large with segment size $w$ = 10, 30, and 60, respectively, and denote them as MSCRED(S), MSCRED(M), and MSCRED(L). Then, we independently evaluate their performances on three types of anomalies, \textit{i.e.}, short, medium, and long with the duration of 10, 30, and 60, respectively. 
The average recall scores over 5 repeated experiments on two datasets are reported in Figure \ref{fig: anomaly_scale_analysis}. We can observe that MSCRED(S) is able to detect all types of anomalies and MSCRED(M) can detect both medium and long duration anomalies. On the contrary, MSCRED(L) can only detect the long duration anomaly. Accordingly, we can interpret the anomaly severity by jointly considering the three anomaly scores. The anomaly is more likely to be long duration if it can be detected in all three channels. Otherwise, it may be a short or medium duration anomaly. 
To better show the effectiveness of MSCRED, Figure \ref{fig: diagnosis-case-study-MCRNN} provides a case study of anomaly diagnosis in power plant data. In this case, MSCRED(S) detects all of 5 anomalies including 3 short, 1 medium and 1 long duration anomalies. MSCRED(M) misses two short duration anomalies and MSCRED(L) only detects the long duration anomaly. Moreover, four residual signature matrices of injected anomaly events show the root causes identification results. We can accurately pinpoint more than half of the anomaly root causes (rows/columns highlighted by red rectangles) in this case. 
\begin{figure}
\begin{center}
\includegraphics[scale=0.4]{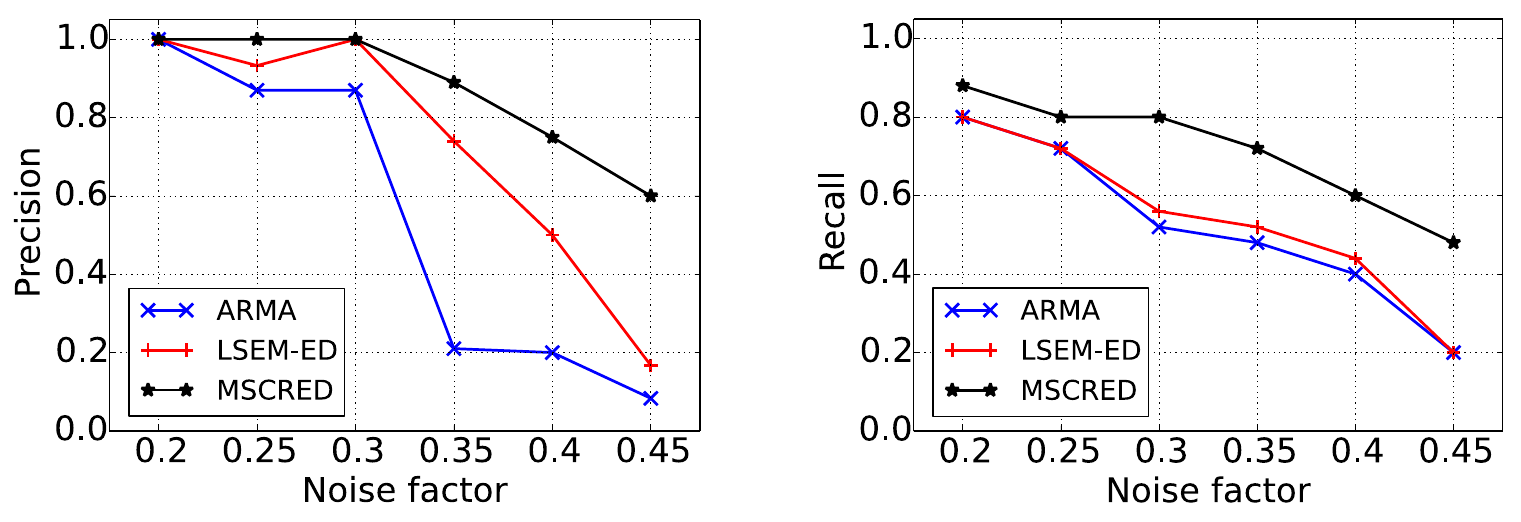}
\vspace{-0.2in}
\caption{Impact of data noise on anomaly detection.}
\label{fig: noise_analysis}
\vspace{-0.3in}
\end{center}
\end{figure}
\subsubsection{Robustness to Noise (RQ5).}
The multivariate time series often contains noise in real world applications, thus it is important for an anomaly detection algorithm to be robust to input noise. To study the robustness of MSCRED for anomaly detection, we conduct experiments in different synthetic datasets by adding various noise factors $\lambda$ in Equation 7. Figure \ref{fig: noise_analysis} shows the impact of $\lambda$ on the performance of MSCRED, ARMA, and LSTM-ED. Similar to previous evaluation, we compute Precision and Recall scores based on the optimized cutting threshold and the average values of 5 repeated experiments are reported for comparison. We can observe that MSCRED consistently outperforms ARMA and LSTM-ED when the scale of noise varies from 0.2 to 0.45. This suggests that, compared with ARMA and LSTM-ED, MSCRED is more robust to the input noise. 

\vspace{-1mm}
\section{Conclusion} \label{sec:conclusion}
In this paper, we formulated anomaly detection and diagnosis problem, and developed an innovative model, \textit{i.e.}, MSCRED, to solve it. MSCRED employs multi-scale (resolution) system signature matrices to characterize the whole system statuses at different time segments and adopts a deep encoder-decoder framework to generate reconstructed signature matrices. The framework is able to model both inter-sensor correlations and temporal dependencies of multivariate time series. The residual signature matrices are further utilized to detect and diagnose anomalies. Extensive empirical studies on a synthetic dataset as well as a power plant dataset demonstrated that MSCRED can outperform state-of-the-art baseline methods. 
\vspace{-1mm}
\section*{Acknowledgments}
Chuxu Zhang and Nitesh V. Chawla are supported by the Army Research Laboratory under Cooperative Agreement Number W911NF-09-2-0053 and the National Science Foundation (NSF) grant IIS-1447795.
\bibliographystyle{aaai}
\footnotesize
\bibliography{paper}

\end{document}